\documentclass[wcp]{jmlr}
\usepackage{amsmath,amssymb,graphicx,url}
\jmlrvolume{204}
\jmlryear{2023}
\jmlrworkshop{Conformal and Probabilistic Prediction with Applications}
\jmlrproceedings{PMLR}{Proceedings of Machine Learning Research}

\usepackage{algorithm}
\usepackage{algorithmic}

\usepackage{array}
\usepackage{booktabs}   
\usepackage{listings}
\usepackage{longtable}

\newtheorem{assumption}{Assumption}

\title{On training locally adaptive CP}

\author{\Name{Nicolo Colombo}\Email{nicolo.colombo@rhul.ac.uk}\\
\addr{Royal Holloway, University of London, Egham, Surrey, UK}}
\editor{Harris Papadopoulos, Khuong An Nguyen, Henrik Boström and Lars Carlsson}

\begin{document}
\maketitle
\begin{abstract}
We address the problem of making Conformal Prediction (CP) intervals locally adaptive.
Most existing methods focus on approximating the object-conditional validity of the intervals by partitioning or re-weighting the calibration set. 
Our strategy is new and conceptually different. 
Instead of re-weighting the calibration data, we redefine the conformity measure through a trainable change of variables, $A \to \phi_X(A)$, that depends explicitly on the object attributes, $X$.
Under certain conditions and if $\phi_X$ is monotonic in $A$ for any $X$, the transformations produce prediction intervals that are guaranteed to be marginally valid and have $X$-dependent sizes.
We describe how to parameterize and train $\phi_X$ to maximize the interval efficiency.
Contrary to other CP-aware training methods, the objective function is smooth and can be minimized through standard gradient methods without approximations.
\end{abstract}


\section{Introduction}
Two features of Conformal Prediction (CP) contribute to their increasing popularity, finite-sample validity and straightforward applicability.
Unlike the Bayesian approach, CP algorithms are computationally cheap, provide practical non-asymptotic guarantees, and can augment any pre-existing or pre-trained point-prediction model.
The very reasons for their success may hide a few structural limitations.
On the one hand, finite-sample validity is hard to achieve on heteroskedastic data, as the prediction intervals should have attribute-dependent sizes.
On the other hand, defining the same (fixed) CP algorithm for any point-prediction model may be a sub-optimal strategy for certain types of tasks or data. 
Lastly, it is unlikely that machine learning algorithms trained to generate good point-like predictions are automatically CP-efficient without being retrained. 

We introduce a new CP-aware fine-tuning scheme to make CP algorithms more efficient and locally adaptive.
Compared to analogous strategies, our approach does not break data exchangeability.
This implies the obtained locally-adaptive prediction intervals are marginally valid, as the non-adaptive ones, but have attribute-dependent sizes.
Moreover, since the scheme does not modify the underlying point-prediction model, it can be merged with other localization methods.
The idea is to boost the local adaptivity of the prediction intervals by optimizing a set of learnable transformations of the conformity scores.
The trained transformations are guaranteed to produce marginally valid prediction sets by construction. 
The prediction sets become locally adaptive when we map them back to the label space by inverting the trained transformations.

The scheme generalizes the Error Re-weighted Conformal approach (ERC) of \cite{papadopoulos2008normalized, papadopoulos2011regression}, in which the conformity scores are re-weighted by a pre-fitted model of the residuals.
More precisely, let $f$ be a pre-trained point-prediction algorithm, $f(X) \approx {\rm E}_{Y|X}(Y)$, $A$ the standard conformity score, $A = (f(X) - Y)^2$, and $(1-\alpha)\in [0, 1]$ a user-defined confidence level.
ERC can be viewed as a coordinate transformation $A \to \phi_X(A) = \frac{A}{g(X)^2 + \gamma}$, where $g$ is a model of the conditional prediction error, i.e. $g(X) \approx {\rm E}_{Y|X}(f(X) - Y)$, and $\gamma > 0$. 
The boundaries of the prediction intervals are the solutions of $\phi_X(A) = \frac{(Y - f(X))^2}{g(X)^2 + \gamma} = \hat q$, where $\hat q$ is the $(1 - \alpha)$-quantile of the conformity scores distribution, estimated on the calibration set. 
By definition of empirical quantile, the prediction interval at $X$, $C = \{y: \phi_X(A(X, y)) \leq \hat q\}$, is marginally valid.
The size of $C$ is $|C|=\sqrt{\hat q (g(X)^2 + \gamma)}$ and explicitly depends on $X$, which means the intervals grow and shrink over the attribute space.

We extend the ERC idea in two ways, \footnote{For simplicity, we focus on regression and make standard i.i.d. assumptions, but the approach generalizes with minor changes to classification problems and exchangeable data.}  
\begin{itemize}
    \item we define a general parametric class of coordinate transformations, $\Phi_\theta $, that are guaranteed to produce marginally valid and locally adaptive prediction intervals (Sections \ref{section coordinate transformation}, \ref{section model classes} and \ref{section validity of the prediction intervals}) and    
    \item we show how to find a $\Phi_\theta$ that maximizes the efficiency of the intervals (Sections \ref{section efficiency of the prediction intervals} and \ref{section gradient optimization}).
\end{itemize}
To guarantee the validity of the transformed prediction sets and their interpretability as prediction intervals in the label space, we require the attribute-dependent transformations, $\phi_X \in \Phi_\theta$, to be monotonic in $A$ and have the same codomain for all $X$ (Assumption \ref{assumption phi}). 
Validity and local adaptability are proven in Theorem \ref{theorem validity} by adapting a standard proof of CP marginal validity and using the assumed invertibility of $\phi_X$. 
To train $\Phi_\theta$, we introduce a confidence-specific cost function that measures the non-efficiency of the intervals at a given confidence level ($\ell_\alpha$ in \eqref{single alpha objective}).
The confidence-specific cost function contains non-differentiable terms but becomes smooth and tractable when it is averaged over all possible confidence levels and estimated empirically on a finite-size data set ($\ell_{all}$ in \eqref{all alphas objective expectation}).
This makes the learning task a smooth and unconstrained optimization problem, which can be solved with gradient approaches even if the inverse map, $\phi_X^{-1}$, is not available analytically.
In the experiments, we compare four classes of attribute-dependent transformations with the standard non-adaptive CP algorithm and ERC.

\subsection{Related work}
\label{section related work}

Redefining the conformity functions to include an attribute-dependent factor is not new. 
The first example of this technique is the ERC method of \cite{papadopoulos2008normalized, papadopoulos2011regression} (see also Section 5 of \cite{romano2019conformalized}).
Our transformation of the conformity score is more general and is not pre-trained by fitting the residuals.

A growing stream of works addresses the problem of producing prediction intervals that are approximately object-conditional valid (whit finite-size data).
Approximate conditional validity is obtained through an attribute-dependent re-weighting of the calibration samples. 
Data exchangeability is temporarily broken and needs to be restored at the end.
The idea of \cite{lei2014distribution} and \cite{vovk2012conditional} has been refined in the last few years by establishing links to Kernel Density Estimation and the covariate shift problem (\cite{barber2022conformal, han2022split, guan2023localized}).
Here, we follow a conceptually different path, where data exchangeability is preserved at all stages.

In the Conformalized Quatile Regression model of \cite{romano2019conformalized}, the standard conformity function is replaced with a proxy of the pinball loss.
Given $(1 - \alpha) \in [0, 1]$, the optimum of the pinball loss is an estimate of the conditional $(1-\alpha)$-quantile function.
This implies the method can not quantify the uncertainty of a pre-trained point-prediction model and needs to be retrained for each $\alpha \in [0, 1]$. 
See \cite{sesia2020comparison} for a review of conformal quantile regression methods.

Evaluation scores similar to $\ell_{\alpha}$ defined in \eqref{single alpha objective} are popular in the CP literature.
Recent works exploit objective functions similar to \eqref{single alpha objective} to train the underlying prediction models in a CP-aware way. 
Here, we fix the underlying prediction model and learn a new conformity function, $\phi_{X}(A)$.
We also average $\ell_\alpha$ over $\alpha \in [0, 1]$, which makes the optimization problem smooth.
This is a significant improvement compared to most CP-aware objectives used in the past, which often require smooth relaxations that may be hard to optimize (\cite{bellotti2020constructing, stutz2021learning, einbinder2022training}).  

A recent work, \cite{einbinder2022training}, is particularly close to ours.
The authors propose to train a point-prediction model by forcing the distribution of conformity scores to be uniform on $[0, 1]$ for all $X$.
The definition of the conformity scores does not change.
It would be interesting to investigate whether tuning the underlying model or the conformity score is equivalent, theoretically and practically.
Also, while our approach extends straightforwardly to classification tasks (provided the distribution of the conformity scores is continuous), it may be hard to adapt \cite{einbinder2022training} to the regression setup.
As for Conformalized Quantile Regression, the CP algorithm of \cite{einbinder2022training} requires fine-tuning the underlying model and cannot quantify the uncertainty of a pre-trained non-conformal classifier.

Using the Inverse Function Theorem to compute the gradient of implicit functions is a common technique for minimizing objective functions without analytical form (\cite{krantz2002implicit}). 
The idea has become popular in the domain of implicit layers optimization (\cite{duvenaud2020deep}).

\subsection{Contribution}
To the best of our knowledge, this is the first time locally adaptive and efficient conformal prediction intervals are obtained from a data-driven definition of the conformity scores.
Previous works on training CP, e.g. \cite{bellotti2020constructing}, \cite{stutz2021learning}, or \cite{ einbinder2022training}, also aim to optimize the efficiency of the intervals but focus on tuning the underlying point-predictor instead of the CP construction on top of it.
See Section \ref{section related work} for a brief discussion of those works.

Proposing a differentiable CP-aware objective function is also relevant. 
The averaged objective function presented here may help define a differentiable version of \cite{bellotti2020constructing}, \cite{stutz2021learning}, or \cite{einbinder2022training}.

As \cite{papadopoulos2008normalized}, our approach is orthogonal to methods that fine-tune the underlying point-prediction model, e.g. \cite{romano2019conformalized}, and is compatible with localization approaches based on the selection/re-weighting of the calibration objects (\cite{lei2014distribution, izbicki2019flexible, romano2020classification, foygel2021limits, sesia2021conformal, guan2023localized}).
Merging different localization techniques, as in \cite{guan2023localized}, may boost the usability of CP on specific tasks.

Finally, we believe this work is one of the first to exploit, at the same time, the invariance of CP intervals under monotone transformations of the conformity scores and recent ideas on conformity-aware training.

\section{Methods}
\label{section methods}
\subsection{A regression task}
\label{section regression task}
Let $\{D, (X_{test}, Y_{test})\} = \{(X_n, Y_n) \in {\cal X} \times {\mathbb R}\}_{n=1}^{N} \cup (X_{test}, Y_{test})$ be a collection of i.i.d. random variables following an unknown joint distribution, $P_{XY}$,  
$f:{\cal X} \to {\mathbb R}$ a pre-trained point-prediction model that approximates the object-conditional mean of $Y$, i.e. $f(X) \approx {\rm E}_{Y|X}(Y)$, and 
\begin{align}
\label{data set}
{\cal D} = \{ (x_n, y_n) \in {\cal X} \times {\mathbb R} \}_{n=1}^{N+1}
\end{align}
a sample of $\{D, (X_{test}, Y_{test})\}$. 
In what follows, we use ${\cal D}$ to train a set of conformity score transformations, $\Phi_\theta$, through a Leave-One-Out strategy where $N$ samples are interpreted as a calibration set and the remaining sample as a test object. 
The resulting CP algorithm will be tested on unseen data, i.e. a new sample of $\{D, (X_{test}, Y_{test})\}$.

\subsection{Prediction intervals}
\label{section prediction intervals}
Given $f$ and $\{D, (X_{test}, Y_{test})\}$, CP algorithms use a conformity function, $a = a(f(X), Y)$, to evaluate the conformity of $f(X_{test})$ with the predictions made on the calibration set, $\{ f(X_{n})\}_{n=1}^N$.
Let  $(1-\alpha) \in [0, 1]$ be a user-defined confidence level.
The corresponding prediction set
\begin{align}
\label{prediction interval general}
C = \{ y: a(f(X_{test}, y)) \leq \hat q\}, \quad \hat q \ {\rm s.t.} \left| \{ a(f(X_n), Y_n) \leq \hat q \}_{n=1}^N \right| = \lceil (N + 1) (1 - \alpha) \rceil 
\end{align}
where $|{\cal S}|$ is the cardinality of ${\cal S}$, is said to be marginally valid with confidence $(1-\alpha)$ because 
\begin{align}
    {\rm Prob}\left( Y_{test} \in C\right) \geq 1 - \alpha
\end{align}
Lower bounds for ${\rm Prob}\left( Y_{test} \in C\right)$ can also be obtained as a function of $N$.   
When $a$ is monotonic in $|f(X) - Y|$, $C$ is a symmetric interval centred on $f(X_{test})$.
The size of $C$ can be associated with the reliability of $f(X_{test})$.
For example, if $a(f(X), Y) = (f(X)-Y)^2$, the prediction interval can be written explicitly as $C = [f(X_{test}) - \Delta, f(X_{test}) + \Delta]$, where $\Delta = 2 a^{-1}(\hat q) = 2 \sqrt{\hat q}$ and $a^{-1}$ is defined by $S = a^{-1}(a(S))$.  
Other possible choices for $a$ include $a_{abs} = |f(X) - Y|$ or $a_{log} = \log((f(X) - Y)^2)$.
Here, we always let $a = (f(X)-Y)^2$. 
Some advantages of using $a_{log}$ are outlined in Section \ref{section examples}. 

\subsection{Coordinate transformations}
\label{section coordinate transformation}
CP intervals are invariant under composition with a monotone function, e.g. if $a \to a_{log} = \log \sqrt{a}$ or $a \to a_{abs} = \sqrt{a}$.
This happens because the intervals only depend on the ranking of the conformity scores, which is preserved if the conformity scores are transformed monotonically.  
This work is a generalization of this idea.
Let $\Phi_\theta = \{\phi_X:{\mathbb R}_+ \to {\cal B}_X, X \in {\cal X}\}$ be a set of attribute-dependent parametric functions of a base conformity scores, $A = a(f(X), Y) \in {\mathbb R}_+$.
The task is to find a $\Phi_\theta$ that maximizes the efficiency of a CP algorithm based on the conformity scores transformed by $\phi_X \in \Phi_\theta$. 
To avoid overfitting, all $\phi_X \in \Phi_\theta$ should share a smooth functional dependence on $X$, e.g. we may require $\phi_X(A) \sim \phi_{X'}(A)$ if $X \sim X'$ for all $A \in {\mathbb R}_+$.
Some assumptions are needed to guarantee that $B=\phi_X(A)$ is a well-defined conformity score. 
If so, the transformed prediction sets are defined like in \ref{prediction interval general} for any $A$ and $X$.
More explicitly, we let 
\begin{align}
\label{prediction interval transformed implicit}
C = \{ y: \phi_{X_{test}}(A), y))\leq \hat q\}, \quad \hat q \ {\rm s.t.} \left| \{ \phi_{X_n}(A_{n}) \leq \hat q \}_{n=1}^N \right| = \lceil (N + 1) (1 - \alpha) \rceil 
\end{align}
where $(X_n, Y_n)\in D$, $\phi_{X_n} \in \Phi_\theta$, $A_n = a(f(X_n), Y_n)$, $n=1, \dots, N$, and $\alpha \in [0, 1]$.
If $a(f(X), Y) = (f(X) - Y)^2$ and under certain assumptions on $\Phi_\theta$, e.g. if $\Phi_\theta$ is defined as in Assumption \ref{assumption phi} (see below), $C$ can be rewritten as  
\begin{align}
\label{prediction interval transformed explicit}
C = [f(X_{test}) - \Delta, f(X_{test}) + \Delta], 
\quad 
\Delta = \sqrt{\phi^{-1}_{X_{test}}(\hat q)} 
\end{align}
Differently from the case of a rigid, i.e. object-independent, monotone transformation, the prediction intervals produced by different $\Phi_\theta$ may not be equivalent.
See Section \ref{section examples} for an example.
\subsection{Model classes}
\label{section model classes}
In the examples of Section \ref{section examples}, the coordinate transformations belong to simple model classes, e.g. $ \Phi_\theta = \{\phi_X(A) = \sqrt{A} - \theta X, \theta \in {\mathbb R}\}$.
Varying the value of the parameter with a class produces non-equivalent intervals. 
Choosing different classes may also change the intervals.
For concreteness, we restrict ourselves to model classes that satisfy the following assumption. 
\begin{assumption}
\label{assumption phi}
Let ${\cal X} = {\mathbb R}^d$, ${\cal B}_X \subseteq {\mathbb R}$, and 
\begin{align}
\Phi_\theta = \{ \phi_X: {\mathbb R}_+ \to {\cal B}_X, X \in {\cal X} \}
\end{align}
$\Phi_\theta$ is such that
\begin{align}
&\phi_{X}'(A)  > 0, \quad {\rm for \ all \ } X \in {\cal X} \ {\rm and \ all \ } A \in {\mathbb R}_+ \\
&{\cal B}_X = {\cal B}_{X'}, \ {\rm for \ all \ } X, X' \in {\cal X}
\end{align}    
where $A = a(f(X), Y)$, $a = (f(X) - Y)^2$,  and $h'(s) = \frac{d}{ds} h(s) =  \frac{d}{ds'} h(s')|_{s' = s}$.
\end{assumption}
Requiring $\phi_{X}' >0$ guarantees that $B=\phi_X(A)$ is a well-defined conformity score and the existence of the inverse map, $\phi_{X}^{-1}: {\cal B}_X \to {\mathbb R}_+$, which is defined implicitly by $\phi_{X}^{-1}(\phi_{X}(A)) = A$.
Requiring the codomain of the transformations to be the same for all $X$ guarantees that we can convert the transformed prediction sets in \eqref{prediction interval transformed implicit} into lable-space symmetric intervals centred in $f(X_{test})$.
Imagine that $\Phi_\theta$ does not fulfil the second requirement of Assumption \ref{assumption phi} and, for example, ${\cal B}_{test} \subset {\cal B}_{X_{n}}$ for any $n=1, \dots, N$.
Then the inequalities in \eqref{prediction interval transformed implicit} may not have a real solution because $\hat q = \phi_{X_{n*}}(A_{n*})$, $n* \in \{1, \dots, N \}$, may lay outside the codomain of $\phi_{X_{test}}$. Both assumptions are satisfied if, for example, $\phi_X(A) \to \phi_X(\log A) + \epsilon \log A $, $\phi_X'(A) > 0 $ for all $X$ and $A$, and $\epsilon > 0$.

\subsection{Examples}
\label{section examples}
\subsubsection{Inequivalent intervals}
\label{section inequivalent intervals}
Let ${\cal X} = {\mathbb R}$, $B = \phi_X(A) = \sqrt{A} - \theta X$, and $\theta_\pm =\pm 1$.
Let the calibration set be $\{ (x_n, y_n)\}_{n=1}^3$, $f$ such that $(a_1, x_1) = (1, 1)$,  $(a_2, x_2) = (2, 2)$, and $(a_3, x_3) = (3, 3)$, $x_{test} = 0$, and $\alpha = \frac12$. 
If $\theta = 1$, $\{ b_n\}_{n=1}^3 = \{2, \sqrt{2}  + 2, \sqrt{3} + 3\}$ and $\hat q = \hat q_+ = \sqrt{2} + 2$.
If $\theta = -1$, $\{ b_n\}_{n=1}^3 = \{0, \sqrt{2}  - 2, \sqrt{3} - 3\}$ and $\hat q = \hat q_- = \sqrt{2} - 2$. 
The prediction intervals, $C_{\pm} = [f(x_{test}) - \Delta_\pm, f(x_{test}) + \Delta_\pm ]$ with $\Delta_\pm = \sqrt{\phi_{x_{test}}^{-1}(\hat q_\pm)} = \sqrt{(\hat q_\pm + \theta x_{test})^2} = |\hat q_\pm|$, have sizes $|C_+| = 2( 2 + \sqrt{2}) > |C_-| = 2(2-\sqrt{2})$, i.e. $C_+$ and $C_-$ have different sizes.

\subsubsection{ERC}
\label{section erc}
Let $a = (f(X) - Y)^2$ and $\phi_X$ be the ERC coordinate transformations of \cite{papadopoulos2008normalized}, i.e.
\begin{align}
\phi_{X}(A) = \frac{A}{g(X)^2 + \gamma},\quad A = (f(X) - Y)^2, \quad g:{\mathbb R}^d \to {\mathbb R}, \quad \gamma > 0 
\end{align}
$\Phi_\theta = \{ \phi_X: {\mathbb R}_+ \to {\cal B}_X, X \in {\cal X}\}$ fulfils the requirements of Assumption \ref{assumption phi} because $\phi'_X(A) = \frac{1}{g(X)^2 + \gamma} > 0$ and ${\cal B}_X = {\mathbb R}_+$ for all $X$.
The inverse transformation is $ \phi^{-1}_X(B) = B (g(X)^2 + \gamma)$ and $C = [f(X_{test}) - \Delta, f(X_{test}) + \Delta]$, where $\Delta = \sqrt{\hat q (g(X_{test})^2 + \gamma)}$ and $\hat q$ is the transformed conformity score of a calibration object.
If $\hat q $ is the $(\lceil(N + 1)(1 - \alpha)\rceil)$th smallest object of $\{ (\phi_{X_n}(A_n)) \}_{n=1}^N$, the interval is marginally valid with confidence $(1 - \alpha)$ and has $X_{test}$-dependent size $|C|=2\Delta$.
In \cite{papadopoulos2008normalized}, $g$ is chosen to be a model of the conditional residuals, i.e. $g(X) \approx {\rm E}_{Y|X}(f(X) - Y)$, but other choices are also possible. 

\subsubsection{Non-adaptive transformations}
\label{section non adaptive transformations}
If the transformation does not depend on $X$, the second requirement in Assumption \ref{assumption phi} is always satisfied. 
For example, consider
\begin{align}
\phi_{0}(A) = \log A + \gamma, \quad A = (f(X) - Y)^2, \quad \gamma \in {\mathbb R}
\end{align}
then $\phi_0'(A) = \frac{1}{A} > 0$, ${\cal B}_X = (-\infty, \infty)$ for all $X$, and $\phi_0^{-1}(B) = \exp{(B - \gamma)}$. The corresponding prediction intervals have size $|C| = 2 \sqrt{\exp{(\hat q - \gamma)}}$, which does not depend on $X_{test}$.

\subsubsection{Different codomains}
\label{section different codomains}
A set of transformations that satisfy $\phi_X'(A)>0$ for all $X \in {\cal X}$ but may have ${\cal B}_X \neq {\cal B}_{X'}$ is
\begin{align}
\phi_{X}(A) = A + g(X)^2 ,\quad A = (f(X) - Y)^2, \quad g:{\mathbb R}^d \to {\mathbb R} 
\end{align}
as $\phi_X'(A)  = 1 > 0$ and ${\cal B}_{X} = [g(X)^2, \infty) \subset {\mathbb R}_+$ for all $X$ and $A$. 
The inverse transformation, $\phi_X^{-1}(B) = B - g(X)^2$ is well defined for all $B$ and all $X$.
The prediction intervals, however, may not have a straightforward interpretation in the label space.
If $\hat q = \phi_{X_{n*}}(A_{n*}) = A_{n*} + g(X_{n*})$ is the $(\lceil(N + 1)(1 - \alpha)\rceil)$=th smallest object of $\{B_n = A_n + g(X_n)^2 \}_{n=1}^N$, the interval at $X_{test}$ is defined by $A_{test} \leq A_{n*} + g(X_{n*})^2 - g(X_{test})^2$, which is meaningless if $A_{n*} + g(X_{n*})^2 < g(X_{test})^2$ because $A_{test} = (f(X_{test})-Y_{test})^2 > 0$. 
The problem does not arise if $\phi(A) \to  \phi(\log A)$.

\subsection{Validity of the prediction intervals}
\label{section validity of the prediction intervals}
Global monotone transformations of the conformity scores do not affect the resulting prediction intervals. 
The first example of Section \ref{section examples} shows this is not the case for locally-defined transformations.
Under Assumption \ref{assumption phi}, however, the inverse map produces label-space prediction intervals that are marginally valid and have attribute-dependent sizes.
\begin{theorem}
\label{theorem validity}
    Let $(X_1, Y_1), \dots, (X_N, Y_N), (X_{test}, Y_{test}) \sim P_{XY}$ be a collection of i.i.d. random variables, 
    $f \sim {\rm E}_{Y|X}(Y)$ a point-prediction model, 
    and $A_n = (f(X_n) - Y_n)^2$, $n = 1, \dots, N$.
    Let $\Phi_\theta$ be a family of strictly increasing scalar functions satisfying the requirements of Assumption \ref{assumption phi} and 
    $B_n = \phi_{X_n}(A_n)$, $n= 1, \dots, N$.
    Assume $B_n \neq B_{n'}$ if $n \neq n'$.
    Then there exists a permutation of $\{1, \dots, N \}$, $\{ m_n = \pi(n) \}_{n=1}^N$  such that $ B_{m_1} < \dots < B_{m_n} <\dots < B_{m_N}$ and, for any $\alpha \in [\frac{1}{N+1}, 1]$,
    the interval 
    \begin{align}
    \label{prediction interval theorem}
        C = [f(X_{test}) - \Delta, f(X_{test}) + \Delta],
        \quad \Delta = \sqrt{\phi^{-1}_{X_{test}}\left( B_{m*})\right)}, 
        \quad m* = \lceil (N + 1)(1 - \alpha)\rceil
    \end{align}
    is marginally valid with confidence $1 - \alpha$, i.e.
    \begin{align}
        {\rm Prob}\left( Y_{test} \in C)\right) \geq 1 - \alpha 
    \end{align}
\end{theorem}

\paragraph{Proof of Theorem \ref{theorem validity}}
If $B_n \neq B_{n'}$ if $n \neq n'$, the exchangeability of $(X_n, Y_n)$, $n-1, \dots, N$, and $(X_{test}, Y_{test})$ implies 
\begin{align}
    {\rm Prob}(B_{test} \leq B_{m_n}) = \frac{m_n}{N + 1}, \quad n=1, \dots, N
\end{align}
where $B_{test} = \phi_{X_{test}}(A_{test}) = \phi_{X_{test}}((f(X_{test}) - Y_{test})^2)$.
In particular, 
\begin{align}
    {\rm Prob}(B_{test} \leq B_{m*}) = \frac{\lceil (N + 1)(1 - \alpha)\rceil}{N + 1} \geq 1 - \alpha
\end{align}
or, equivalently, ${\rm Prob}(\phi_{X_{test}}(A_{test}) \leq B_{m*}) \geq 1 - \alpha$.
The strict monotonicity of $\phi_X(A)$ for all $X \in {\mathbb R}^d$ implies that 
$\phi_{X_{test}}$ is invertible with
inverse $\phi^{-1}_{X_{test}}: {\cal B}_{X_{test}} \to {\mathbb R}_+$ defined by 
$A = \phi_{X_{test}}^{-1}(\phi_{X_{test}}(A))$.
$\phi^{-1}_{X_{test}}$ and is also strictly increasing because  
$\phi_{X_{test}}^{-1'} = \frac{1}{\phi_{X_{test}}'}$ and $\phi_{X_{test}}'>0$. \footnote{$\phi_{X_{test}}^{-1'} = \frac{1}{\phi{X_{test}}'}$ follows from 
$1 = \frac{d}{dA} \phi{X_{test}}^{-1}(\phi{X_{test}}(A)) 
= \phi_X'(\phi{X_{test}}^{-1}(A))\phi_{X_{test}}^{-1'}(A)$.}  
The validity of \eqref{prediction interval theorem} follows from 
\begin{align}
    1 - \alpha &\leq  {\rm Prob}\left(B_{test}\leq B_{m*}\right) \\ 
     &
     = {\rm Prob}\left(\phi^{-1}_{X_{test}}(B_{test}) \leq \phi^{-1}_{X_{test}}(B_{m*})\right) \\
    & = 
    {\rm Prob}\left(A_{test} \leq \phi^{-1}_{X_{test}}(B_{m*})\right) \\
    & =
    {\rm Prob}\left((f(X_{test}) - Y_{test})^2 \leq \phi^{-1}_{X_{test}}(B_{m*})\right) \\
    & =
    {\rm Prob}\left(Y_{test} \in [f(X_{test}) - \Delta, f(X_{test}) + \Delta]\right)
\end{align}
where $\Delta = \sqrt{\phi^{-1}_{X_{test}}(B_{m*})}$ and the first equality holds because $\phi_{X_{test}}^{-1}$ is strictly increasing.  $\square$

\subsection{Efficiency of the prediction intervals}
\label{section efficiency of the prediction intervals}
If the conformity function does not depend on $X$, the size of the prediction intervals is minimal. \footnote{From
the definition of empirical quantile.}
If the conformity scores are locally re-defined, i.e. $A \to \phi_{X}(A)$, the intervals can grow and shrink over the attribute space. 
Under Assumption \ref{assumption phi}, Theorem \ref{theorem validity} ensures that the obtained locally-adaptive intervals remain valid.
Their sizes, $|C| =  2 \sqrt{\phi^{-1}_{X_{test}}(\hat q)}$, which depend on $X_{test}$, $\Phi_\theta$, and the data, are not guaranteed to be optimal.
Exact (optimal) conditionally-valid prediction intervals for all $X$ can be obtained only if a finite-size data set is available for any $X \in {\cal X}$, which is impossible for real-valued attributes (\cite{vovk2012conditional, lei2014distribution}).
Like other locally-adaptive CP algorithms, our scheme aims to approximate such unachievable conditional validity with finite data.
To evaluate the efficiency of $\Phi_\theta$, we compute the average size of the prediction intervals at a user-defined confidence level, $(1 - \alpha) \in [0, 1]$, i.e. 
\begin{align}
\label{single alpha objective}
\ell_\alpha 
= {\rm E}\left( 2 \sqrt{\phi_{X_{test}}^{-1}(\hat q)} \right), \quad 
\hat q \  {\rm \ s.t. \ } \  |\{ \phi_{X_n}(A_n) \leq \hat q \}|  =  \lceil (N + 1)(1 - \alpha) \rceil
\end{align}
where $A_n = a(f(X_n), Y_n)$, $a(f(X), Y) = (f(X), Y)^2 $, and  $(X_n, Y_n) \in D$.
Training $\Phi_\theta$ by minimizing \eqref{single alpha objective} over $\theta$ has two disadvantages.
Firstly, $\ell_\alpha$ depends on the empirical quantile of the transformed conformity scores, i.e. on the ranking of $\{ B_n = \phi_{X_n}(A_n) \}_{n=1}^N$, which needs to be estimated numerically.
Moreover, the dependence of $\hat q$ on $\Phi_\theta$ may be hard to estimate without ad-hoc smooth relaxations. 
A second possible issue of \eqref{single alpha objective} is it requires retraining the model if the target confidence level changes.

A possible way out is to average $\ell_\alpha$ over all possible confidence levels, $(1-\alpha) \in [0, 1]$.
Taking the expectation of \eqref{single alpha objective} over $\alpha$ produces a model that works well for any $\alpha \in [0, 1]$.
More importantly, it bypasses the non-differentiability of \eqref{single alpha objective} without introducing arbitrary relaxations or approximations.
Intuitively, this happens because sorting the transformed conformity scores becomes redundant if we assume a flat prior over $\alpha \in [0, 1]$, i.e. if we let  
\begin{align}
\label{all alphas objective expectation}
\ell_{all} = {\rm E}_{\alpha \sim {\cal U}_{[0, 1]}} \left( \ell_{\alpha}\right) = \int_{[0, 1]} d\alpha \ \ell_{\alpha}     
\end{align}
where $\ell_\alpha$ is defined in \eqref{single alpha objective}.
The number of non-equivalent confidence levels in an empirical version of $\ell_{all}$ coincides with the size of the calibration set, $|D| = N$. 
Each non-equivalent confidence level is associated with one of the transformed conformity scores, $\{B_n = \phi_{X_n}(A_n)\}_{n=1}^N$.
Since the integral becomes a sum, the commutativity of the addends makes the reordering redundant.
The size-$N$ approximation of \eqref{all alphas objective expectation}, 
\begin{align}
    \label{size objective}
    \ell_{all} = {\rm E}\left( \sum_{n=1}^n \sqrt{\phi^{-1}_{X_{test}}\left(\phi_{X_n}(A_n)\right)} \right) 
\end{align}
is then smooth in $\theta$ if $\Phi_\theta$ is smooth in $\theta$ for all $X$ and $A$.
In practice, $\Phi_\theta$ is a fixed set of parametric transformations.
The free parameter, $\theta$, controls the shared functional dependence of all $\phi_X \in \Phi_\theta$.
Specific functional dependencies may be chosen to ensure that $\Phi_\theta$ is differentiable in $\theta$ and satisfies Assumption \ref{assumption phi} (see Section \ref{section experiments} for a few examples). 
Given such $\Phi_\theta$, we can safely estimate \eqref{size objective} and its gradient (see Section \ref{section gradient optimization}) from the training data set, ${\cal D}$ defined in \eqref{data set}, by computing $B_n=\phi_{x_n}((f(x_n)-y_n)^2)$, $n=1, \dots, N$, their empirical quantile, $\hat q$, and $\phi^-1_{x_{test}}(\hat q)$. 
A data-efficient strategy is to average over $N+1$ splits of ${\cal D}$ deformed in \eqref{data set}.
In the $n$-th split, we use $(x_n, y_n)$ as a test object and the remaining objects as a calibration set. 

\subsection{Gradient optimization}
\label{section gradient optimization}
The smoothness of \eqref{size objective} allows using any gradient-based optimization method.
As $\phi_{X}^{-1}(\phi_{X'}(A))$ depends on $\theta$ directly and through its argument, the parameter updates are $\theta \to \theta + \eta d_\theta \ell_{all}$, where $\eta > 0$ and $d_\theta\ell_{all}$ is the total derivative of \eqref{size objective} with respect to $\theta$, i.e.
\begin{align}
    d_\theta \ell_{all}  = {\rm E}\left( \sum_{n=1}^n \frac{
    \nabla_\theta\phi_{X_{test}}^{-1}\left(\phi_{X_n}(A_n)\right) +
    \phi_{X_{test}}^{-1'}\left(\phi_{X_n}(A_n)\right)\nabla_\theta \phi_{X}(A_n)}
    {2\sqrt{\phi^{-1}_{X_{test}}\left(\phi_{X_n}(A_n)\right)}} \right)
\end{align}
Under Assumption \ref{assumption phi}, the monotone transformations $\phi_X \in \Phi_\theta$ are guaranteed to be invertible. 
Assumption \ref{assumption phi}, however, does not guarantee a closed-form expression of $\phi_{X}^{-1}$.
In that case, $\phi_X^{-1}(B)$ should be obtained numerically through standard root-finding methods, e.g. the Bisection method.
As for the backward pass, $\nabla_\theta \phi_X^{-1}$ and $\phi_X^{-1'}$ can be obtained from $\nabla_\theta \phi_X$ and $\phi_X'$ using 
\begin{align}
    0 &= d_\theta \phi_X\left(\phi_X^{-1}(B)\right) = 
    \nabla_\theta \phi_X\left(\phi_X^{-1}(B)\right) + \phi_X'\left(\phi_X^{-1}(B)\right) \nabla_\theta \phi_X^{-1}(B) \\
    1& = \frac{d}{dB} \phi_X\left(\phi_X^{-1}(B)\right) = 
    \phi_X'\left(\phi_X^{-1}(B)\right) \phi_X^{-1'}(B)
\end{align}
Similar techniques are increasingly popular in machine learning.
Especially because they allow differentiating prediction models that include implicitly-defined outputs, e.g. the numerical solutions of an ODE (\cite{duvenaud2020deep}).

\section{Experiments}
\label{section experiments}
We test the feasibility and performance of the proposed scheme by training and comparing the prediction intervals produced by four different CP algorithms.
 Each algorithm is based on one of the following classes of transformations,
 \begin{align}
     &\Phi_{\tt ERC} = \{\phi_X(A) = \frac{A}{g(X)^2 + \gamma }, \gamma >0  \}\\
     &\Phi_{\tt linear} = \{\phi_X(A) = \log{A} + g(X) \}\\
     &\Phi_{\tt exp} = \{\phi_X(A) = A e^{g(X)} \}\\
     &\Phi_{\tt sigma} = \{\phi_X(A) = \sigma\left( \log A + g(X) \right) \}
 \end{align}
 where $g$ is an unconstrained trainable function of $X$.
 In all experiments, we let $g$ be a fully connected ReLu network with five hidden layers of size 100.
 All model classes satisfy Assumption \ref{assumption phi} and are trained by minimizing $\ell_{all}$ defined in \eqref{size objective}.
 The ERC model, $\Phi_{\tt ERC}$ is also trained as suggested in \cite{papadopoulos2008normalized}, i.e. by minimizing ${\rm E}((g(X) - (f(X)-Y)^2)^2)$.
 In the plots and tables, we refer to that setup as {\tt ERC (error fit)}.
 Our baseline is the non-adaptive CP algorithm, $\Phi_0 = \{\phi_X(A) = A \}$. 
 The underlying point-prediction model is the same for all CP algorithms 
 We use a KNN algorithm with the Euclidean distance and cross-validated $K$ and pre-train it on a separate proper-train data set.
The model efficiencies are measured by computing the interval sizes and empirical validities.

We test all models on a synthetic data set (Section \ref{section synthetic data}) and five real-world benchmark data sets (Section \ref{section real data}).
To optimize the parameterized localization network, $g = g(X, \theta)$, over $\theta$, we use a PyTorch implementation of ADAM (\cite{kingma2014adam}), with a fixed learning rate, default parameters, and mini-batches of size 16.
To avoid overfitting, we evaluate the efficiency of the intervals on a small validation set at each epoch and retrain the model up to the one associated with the most efficient intervals.
All data sets are normalized and split into three parts, one to define the KNN point-prediction algorithm, one for training $\Phi_\theta$, and one for testing.

\subsection{Synthetic data}
\label{section synthetic data}
\begin{figure}
    \centering
    \includegraphics[width=.45\textwidth]{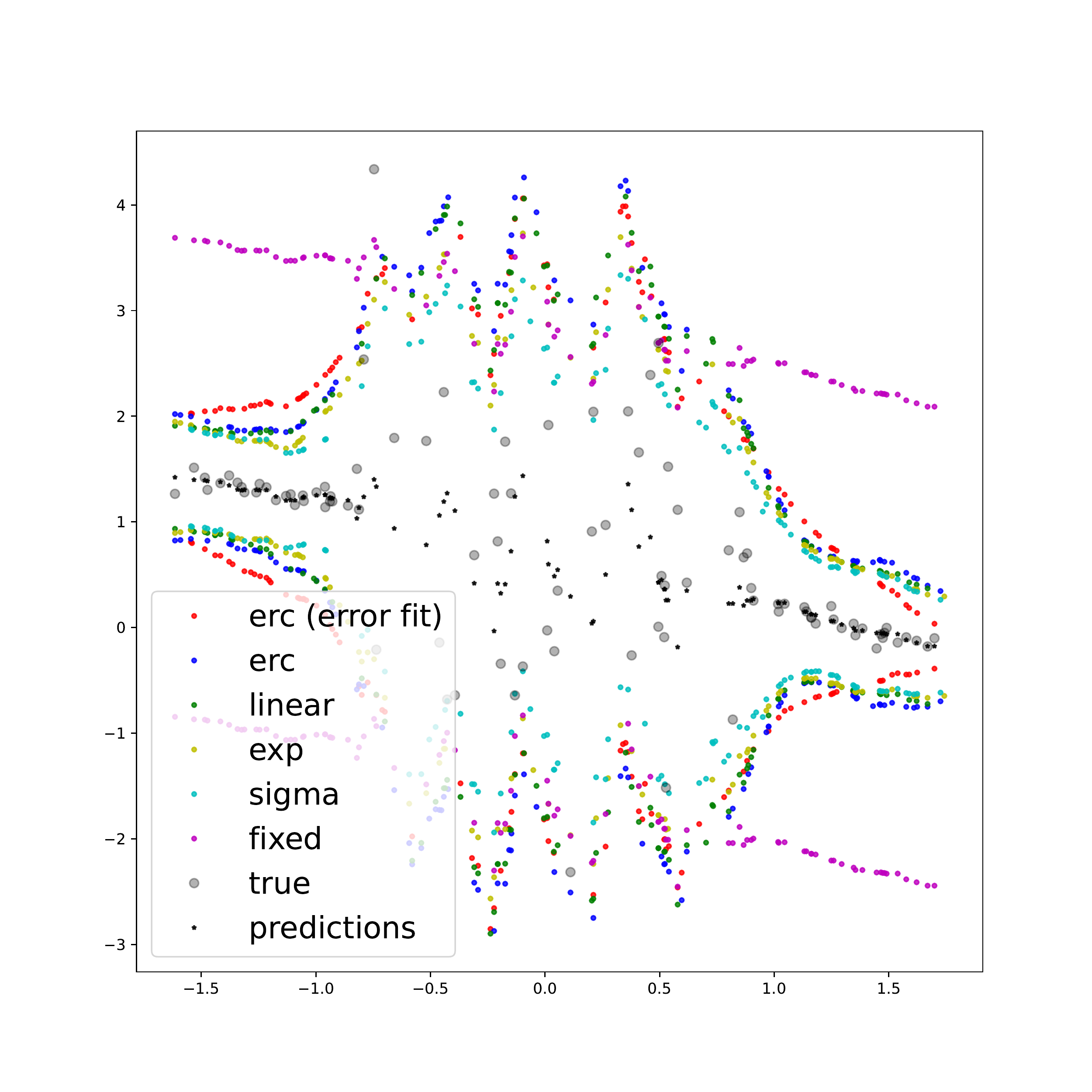}
    \includegraphics[width=.45\textwidth]{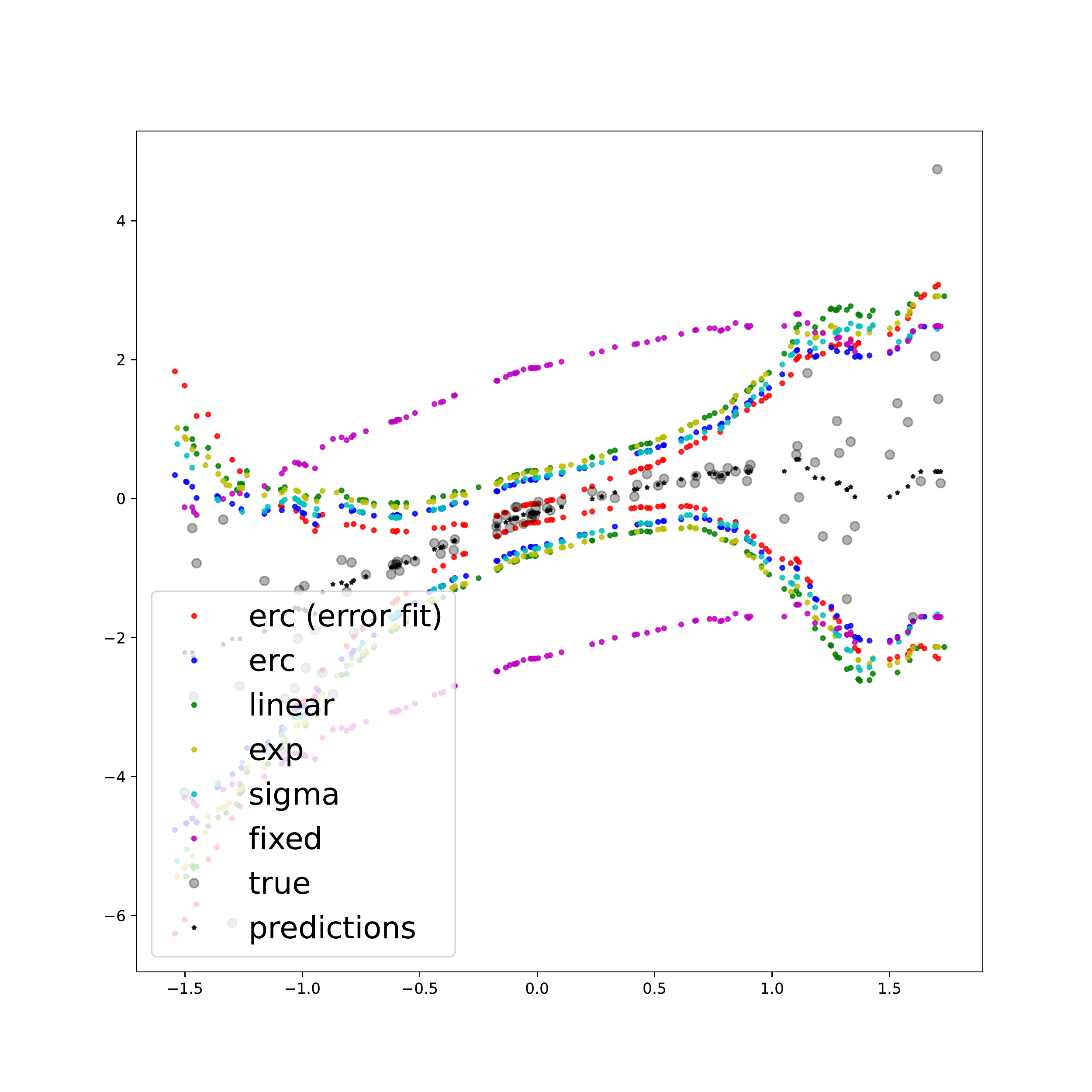}
    \includegraphics[width=.45\textwidth]{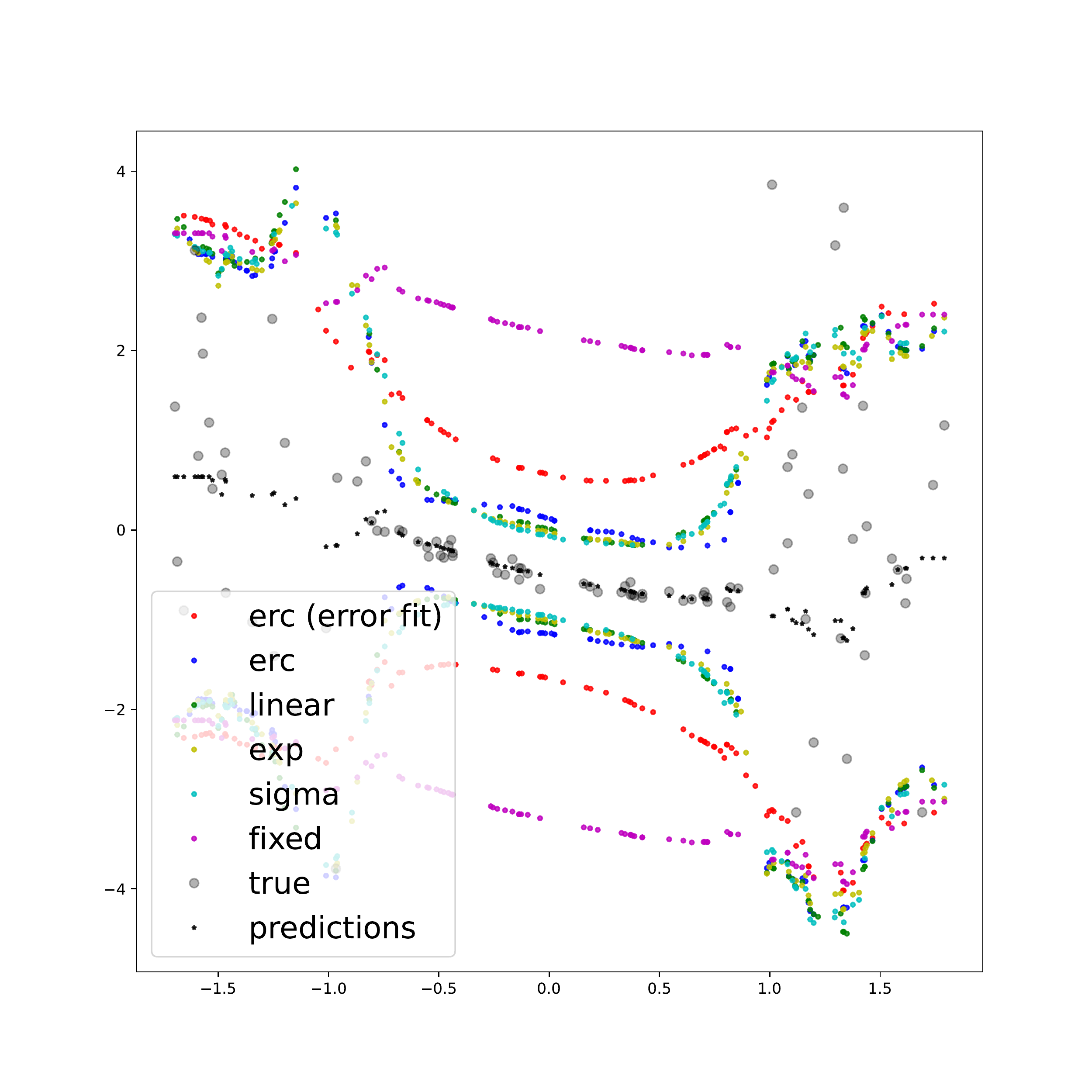}
    \includegraphics[width=.45\textwidth]{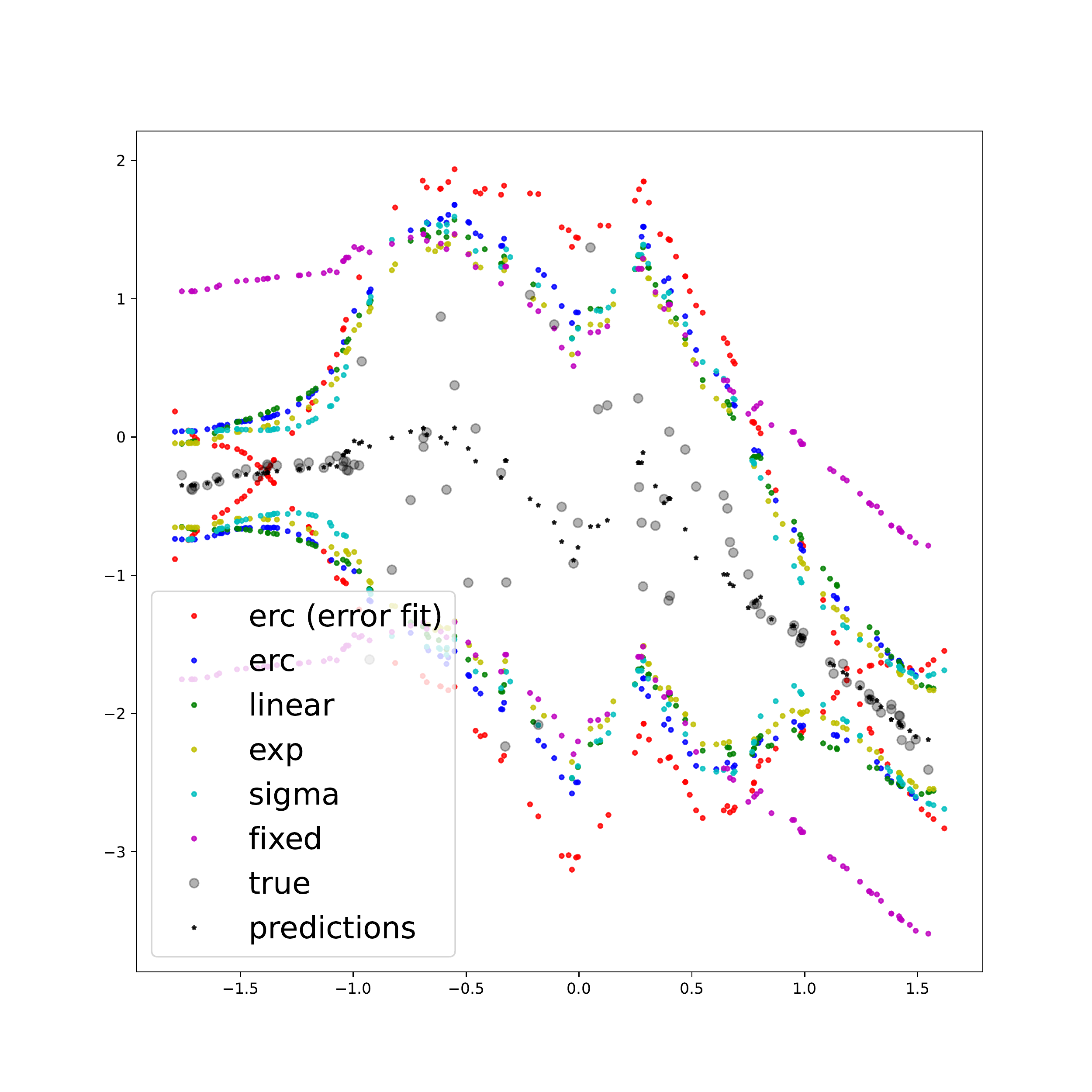}
    \caption{Examples of the locally adaptive intervals obtained with the proposed approach.}
    \label{figure intervals}
\end{figure}

 In Figure \ref{figure intervals}, we show the locally adaptive intervals obtained by the models on four randomly generated synthetic data sets.
 In this case, the confidence level is set to $\alpha=0.05$.
 Each data set consists of 1000 samples generated by perturbing an order-2 polynomial regression model, $f_{true}(X) = w_0 + w_1 X + w_2 X^2$, $w \in {\cal N}(0, 1)^3$, with one of the four following $X$-depedent error functions, 
 \begin{align}
     \varepsilon_{\tt cos} &= \left(\rho + 2 \cos\left(\frac{\pi}{2}|X|\right) \ {\bf 1}(|X|< 0.5)\right) \xi \\
     \varepsilon_{\tt squared} &= \left(\rho + 2 X^2\ {\bf 1}(|X|>0.5) \right) \xi \\
     \varepsilon_{\tt inverse} &= \left(\rho + \frac{2}{\rho + |X|}\ {\bf 1}(|X|>0.5)\right) \xi  \\
     \varepsilon_{\tt linear} &= \left(\rho + (2- |X|)\ {\bf 1}(|X|<0.5)\right) \xi 
 \end{align}
 $\rho = 0.1$, $\xi \sim {\cal N}(0, 1)$.
 In all cases, we sample X uniformly in $[-1, 1]$ and then normalize the input vector $(1, X, X^2)^T$ across all samples. \footnote{Normalizing the input causes the $x$-coordinate of some dots to exceeds $[-1, 1]$ in Figure \ref{figure intervals}.}

Figure \ref{figure intervals} shows that the trained models can produce intervals of locally adaptive size.
As noted in previous works, e.g.  \cite{romano2019conformalized}, {\tt ERC} may become unstable when the localization network, $g$, is trained by fitting the conditional squared errors.
For the fairness of the comparison, we regularize {\tt ERC (error fit)} through the cross-validated early-stopping technique we use for the other models instead of tuning an additional regularization hyper-parameter, $\gamma$.

\subsection{Real data}
\label{section real data}
To check the scalability and efficiency of the algorithms on real-world data comparison, we have used the following regression data sets, 
\begin{itemize}
    \item {\tt energy}, the Energy Efficiency Data Set, 768 observations with eight attributes, \cite{energyData}, 
    \item {\tt concrete}, the Concrete Compressive Strength Data Set, 1030 observations with nine attributes, \cite{concreteData},
    \item {\tt homes}, the King County Home Price Data Set,  21613 observations with 14 attributes, \cite{kcHomeData}, 
    \item {\tt CASP}, the Physicochemical Properties of Protein Tertiary Structure Data Set, 45730 observations with nine attributes, \cite{caspData},
    \item {\tt facebook\_1}, the Facebook Comment Volume Data Set, 40949 observations with 54 attributes, \cite{facebookData}.
\end{itemize} 
These are popular public benchmarks for evaluating the local adaptability of conformal prediction. 
Visit \cite{energyData, concreteData, kcHomeData, caspData, facebookData} or see \cite{romano2019conformalized, sesia2020comparison, guan2023localized} for more details about the data sets. 
Table \ref{table size and val} shows the interval sizes and empirical validities on all datasets and for different confidence levels.
Each value is obtained by averaging the performance over five runs of the training-testing procedure. 

The results confirm that {\tt ERC} may be unstable when the localization function is fitted to the conditional residual.
All models, however, are consistently more efficient than the non-adaptive baseline, $\phi_0$.
The choice of the model class does not look to be crucial, with minor and data set-dependent performance differences.

\begin{table}
\centering
\begin{footnotesize}
\begin{tabular}{*7c}

\toprule \multicolumn{1}{c}{\tt energy }&\multicolumn{2}{c}{$\alpha=0.05$}&\multicolumn{2}{c}{$\alpha=0.1$}&\multicolumn{2}{c}{$\alpha=0.32$}\\ \hline
\midrule&size&val&size&val&size&val\\ 
{\tt erc} (error fit)&$0.86\pm 0.582$&$0.968\pm 0.016$&$0.431\pm 0.289$&$0.924\pm 0.015$&$0.103\pm 0.025$&$0.684\pm 0.066$\\
{\tt erc}&$0.259\pm 0.195$&$0.944\pm 0.023$&$0.184\pm 0.121$&$0.888\pm 0.02$&$0.083\pm 0.017$&$0.684\pm 0.023$\\
{\tt linear}&$0.172\pm 0.018$&$0.948\pm 0.01$&$0.134\pm 0.011$&$0.9\pm 0.028$&$0.078\pm 0.009$&$0.688\pm 0.041$\\
{\tt exp}&$0.161\pm 0.014$&$0.928\pm 0.02$&$0.131\pm 0.007$&$0.868\pm 0.027$&$0.075\pm 0.006$&$0.66\pm 0.046$\\
{\tt sigma}&$0.169\pm 0.027$&$0.948\pm 0.016$&$0.135\pm 0.009$&$0.884\pm 0.008$&$0.078\pm 0.01$&$0.672\pm 0.016$\\
{\tt fixed}&$0.192\pm 0.015$&$0.952\pm 0.01$&$0.143\pm 0.008$&$0.884\pm 0.023$&$0.082\pm 0.008$&$0.66\pm 0.038$\\
\bottomrule
\toprule \multicolumn{1}{c}{\tt concrete }&\multicolumn{2}{c}{$\alpha=0.05$}&\multicolumn{2}{c}{$\alpha=0.1$}&\multicolumn{2}{c}{$\alpha=0.32$}\\ \hline
\midrule&size&val&size&val&size&val\\ 
{\tt erc} (error fit)&$0.736\pm 0.17$&$0.936\pm 0.015$&$0.484\pm 0.072$&$0.888\pm 0.032$&$0.27\pm 0.024$&$0.656\pm 0.015$\\
{\tt erc}&$0.903\pm 0.652$&$0.968\pm 0.016$&$0.637\pm 0.359$&$0.912\pm 0.016$&$0.29\pm 0.049$&$0.672\pm 0.041$\\
{\tt linear}&$0.569\pm 0.088$&$0.96\pm 0.013$&$0.457\pm 0.049$&$0.9\pm 0.018$&$0.267\pm 0.033$&$0.668\pm 0.035$\\
{\tt exp}&$0.567\pm 0.078$&$0.952\pm 0.016$&$0.45\pm 0.032$&$0.92\pm 0.022$&$0.274\pm 0.034$&$0.72\pm 0.025$\\
{\tt sigma}&$0.557\pm 0.073$&$0.952\pm 0.032$&$0.453\pm 0.039$&$0.904\pm 0.015$&$0.268\pm 0.024$&$0.688\pm 0.016$\\
{\tt fixed}&$0.61\pm 0.08$&$0.964\pm 0.02$&$0.472\pm 0.049$&$0.904\pm 0.046$&$0.271\pm 0.025$&$0.688\pm 0.052$\\
\bottomrule
\toprule \multicolumn{1}{c}{\tt homes }&\multicolumn{2}{c}{$\alpha=0.05$}&\multicolumn{2}{c}{$\alpha=0.1$}&\multicolumn{2}{c}{$\alpha=0.32$}\\ \hline
\midrule&size&val&size&val&size&val\\ 
{\tt erc} (error fit)&$1.664\pm 1.165$&$0.936\pm 0.008$&$0.864\pm 0.562$&$0.892\pm 0.01$&$0.234\pm 0.062$&$0.68\pm 0.031$\\
{\tt erc}&$0.526\pm 0.038$&$0.936\pm 0.023$&$0.409\pm 0.047$&$0.876\pm 0.023$&$0.195\pm 0.017$&$0.656\pm 0.027$\\
{\tt linear}&$0.522\pm 0.065$&$0.944\pm 0.015$&$0.425\pm 0.05$&$0.904\pm 0.015$&$0.206\pm 0.024$&$0.704\pm 0.06$\\
{\tt exp}&$0.521\pm 0.054$&$0.94\pm 0.022$&$0.392\pm 0.053$&$0.892\pm 0.01$&$0.197\pm 0.018$&$0.68\pm 0.025$\\
{\tt sigma}&$0.533\pm 0.077$&$0.936\pm 0.02$&$0.421\pm 0.058$&$0.892\pm 0.027$&$0.203\pm 0.027$&$0.632\pm 0.045$\\
{\tt fixed}&$0.721\pm 0.087$&$0.94\pm 0.025$&$0.484\pm 0.093$&$0.896\pm 0.027$&$0.192\pm 0.024$&$0.672\pm 0.037$\\
\bottomrule

\toprule \multicolumn{1}{c}{\tt CASP }&\multicolumn{2}{c}{$\alpha=0.05$}&\multicolumn{2}{c}{$\alpha=0.1$}&\multicolumn{2}{c}{$\alpha=0.32$}\\ \hline
\midrule&size&val&size&val&size&val\\ 
{\tt erc} (error fit)&$1.375\pm 0.093$&$0.948\pm 0.02$&$1.073\pm 0.058$&$0.904\pm 0.023$&$0.643\pm 0.058$&$0.664\pm 0.056$\\
{\tt erc}&$1.412\pm 0.191$&$0.936\pm 0.027$&$1.045\pm 0.048$&$0.888\pm 0.037$&$0.637\pm 0.056$&$0.648\pm 0.07$\\
{\tt linear}&$1.392\pm 0.203$&$0.96\pm 0.018$&$1.018\pm 0.053$&$0.916\pm 0.015$&$0.631\pm 0.07$&$0.684\pm 0.032$\\
{\tt exp}&$1.363\pm 0.187$&$0.944\pm 0.023$&$1.014\pm 0.07$&$0.892\pm 0.037$&$0.635\pm 0.042$&$0.68\pm 0.022$\\
{\tt sigma}&$1.321\pm 0.207$&$0.956\pm 0.023$&$1.03\pm 0.108$&$0.904\pm 0.034$&$0.629\pm 0.075$&$0.688\pm 0.03$\\
{\tt fixed}&$1.492\pm 0.207$&$0.952\pm 0.02$&$1.093\pm 0.086$&$0.9\pm 0.031$&$0.659\pm 0.063$&$0.668\pm 0.037$\\
\bottomrule

\toprule \multicolumn{1}{c}{\tt facebook\_1 }&\multicolumn{2}{c}{$\alpha=0.05$}&\multicolumn{2}{c}{$\alpha=0.1$}&\multicolumn{2}{c}{$\alpha=0.32$}\\ \hline
\midrule&size&val&size&val&size&val\\ 
{\tt erc} (error fit)&$3.404\pm 1.063$&$0.956\pm 0.015$&$1.473\pm 0.293$&$0.916\pm 0.029$&$0.281\pm 0.062$&$0.692\pm 0.032$\\
{\tt erc}&$3.332\pm 1.36$&$0.948\pm 0.024$&$1.389\pm 0.434$&$0.908\pm 0.027$&$0.282\pm 0.061$&$0.684\pm 0.029$\\
{\tt linear}&$3.091\pm 0.694$&$0.944\pm 0.02$&$1.523\pm 0.353$&$0.904\pm 0.015$&$0.289\pm 0.068$&$0.672\pm 0.03$\\
{\tt exp}&$3.268\pm 1.024$&$0.96\pm 0.022$&$1.534\pm 0.305$&$0.916\pm 0.041$&$0.281\pm 0.049$&$0.708\pm 0.02$\\
{\tt sigma}&$3.088\pm 1.211$&$0.944\pm 0.015$&$1.376\pm 0.439$&$0.9\pm 0.025$&$0.292\pm 0.068$&$0.664\pm 0.05$\\
{\tt fixed}&$3.612\pm 1.243$&$0.948\pm 0.035$&$1.519\pm 0.273$&$0.908\pm 0.035$&$0.276\pm 0.063$&$0.656\pm 0.054$\\
\bottomrule
\end{tabular}
\end{footnotesize}

    \caption{Size (left) and empirical validity (right) of the prediction intervals produced by the models on different datasets and for different confidence levels.}
    \label{table size and val}
\end{table}

\section{Limitations and future work}
We recognize that our work should have included a more extensive real-world validation of the methods. 
Firstly, we only consider a simple point-prediction model and do not run a careful ablation study involving more or less advanced regression models.
While, in many cases, a localization network with five hidden layers of size 100 seems to make $\Phi_\theta \in \{\Phi_{\tt ERC}, \Phi_{\tt linear}, \Phi_{\tt exp}, \Phi_{\tt sigma}\}$ flexible enough, the efficiency of the intervals may be improved further by considering more general model classes.
For example, it may be interesting to train a linear combination of $\Phi_{\tt ERC}$, $\Phi_{\tt linear}$, $\Phi_{\tt exp}$, and $\Phi_{\tt sigma}$ with nonnegative weights.
While the model automatically satisfies Assumption \ref{assumption phi}, its inverse is not available in analytic form.
The parameters of the localization network should then be trained through the implicit differentiation strategy described in Section \ref{section gradient optimization}.

As mentioned in Section \ref{section related work}, the proposed approach is not incompatible with existing methods that approximate object-conditional validity by re-weighting the empirical distribution of the conformity scores, e.g. \cite{vovk2012conditional, han2022split, barber2022conformal, guan2023localized},
Combining a global redefinition of the conformity measure with a sample-specific re-weighting of the calibration distribution may help localize the interval on specific tasks.
Including the proposed strategy in the quantile regression approaches of \cite{sesia2020comparison} would also be possible.

In a follow-up of this work, we will also explore how to interpret $\Phi_\theta$ as a Normalizing Flow between the distribution of the original conformity scores and their transformed versions.
The goal is then to force the joint distribution of the conformity scores and the attributes to factorize, i.e. $\Phi_\theta$ should be trained so that $P_{\phi(A)X} = P_{\phi(A)}P_{X}$, as 
this would make conditional and marginal validity equivalent. 

The code to reproduce the numerical experiments described in Section \ref{section experiments} and possibly updated versions of this manuscript can be found in this \href{https://github.com/nicoloRHUL/onTrainingLocalizedCP}{{\color{blue} Github directory}}.

\section*{Acknowledgements}
We thank the COPA reviewers for their useful comments and questions.
We are grateful to V. Vovk for the inspiring discussions in the preliminary stages of this work.
\bibliography{refs.bib}

\end{document}